\documentclass[letterpaper, 10 pt, conference]{ieeeconf}  

\IEEEoverridecommandlockouts                              

\usepackage{amsmath,amsfonts}
\usepackage{algorithmic}
\usepackage{algorithm}
\usepackage{array}
\usepackage[caption=false,font=normalsize,labelfont=sf,textfont=sf]{subfig}
\usepackage{textcomp}
\usepackage{stfloats}
\usepackage{url}
\usepackage{verbatim}
\usepackage{graphicx}
\usepackage{cite}
\usepackage{caption}

\usepackage{graphicx}
\usepackage{amsmath}
\usepackage{amssymb}
\usepackage{booktabs}

\usepackage{epsfig}
\usepackage{multirow}
\usepackage{tabularx}
\usepackage{adjustbox}
\usepackage{siunitx}
\usepackage{textcomp}

\usepackage{stmaryrd} 
\usepackage{latexsym} 

\usepackage{url}

\usepackage{booktabs}

\newcommand{\ie}{\textit{i.e.}}
\newcommand{\eg}{\textit{e.g.}}

\usepackage{xcolor,colortbl}
\definecolor{Gray1}{gray}{0.85}
\definecolor{Gray2}{gray}{0.65}
\usepackage{pifont}

\makeatletter
\let\NAT@parse\undefined
\makeatother
\usepackage[pagebackref=false,breaklinks=true,letterpaper=true,colorlinks,bookmarks=false]{hyperref}

\usepackage[capitalize]{cleveref}
\crefname{section}{Sec.}{Secs.}
\Crefname{section}{Section}{Sections}
\Crefname{table}{Table}{Tables}
\crefname{table}{Tab.}{Tabs.}

\begin{document}

\title{\bf Bridging Spectral-wise and Multi-spectral Depth Estimation \\ via Geometry-guided Contrastive Learning}

\author{Ukcheol Shin\\
CMU RI\\
{\tt\small ushin@andrew.cmu.edu}
\and
Kyunghyun Lee\\
LG AI Research\\
{\tt\small kyunghyun.lee@lgresearch.ai}
\and
Jean Oh\\
CMU RI\\
{\tt\small hyaejino@andrew.cmu.edu}
}

\twocolumn[{%
\renewcommand\twocolumn[1][]{#1}%
\maketitle
\vspace{-0.2in}
\begin{center}
{
\begin{tabular}{c@{\hskip 0.01\linewidth}c@{\hskip 0.01\linewidth}c@{\hskip 0.01\linewidth}c}
\multicolumn{2}{c}{\includegraphics[width=0.44\linewidth]{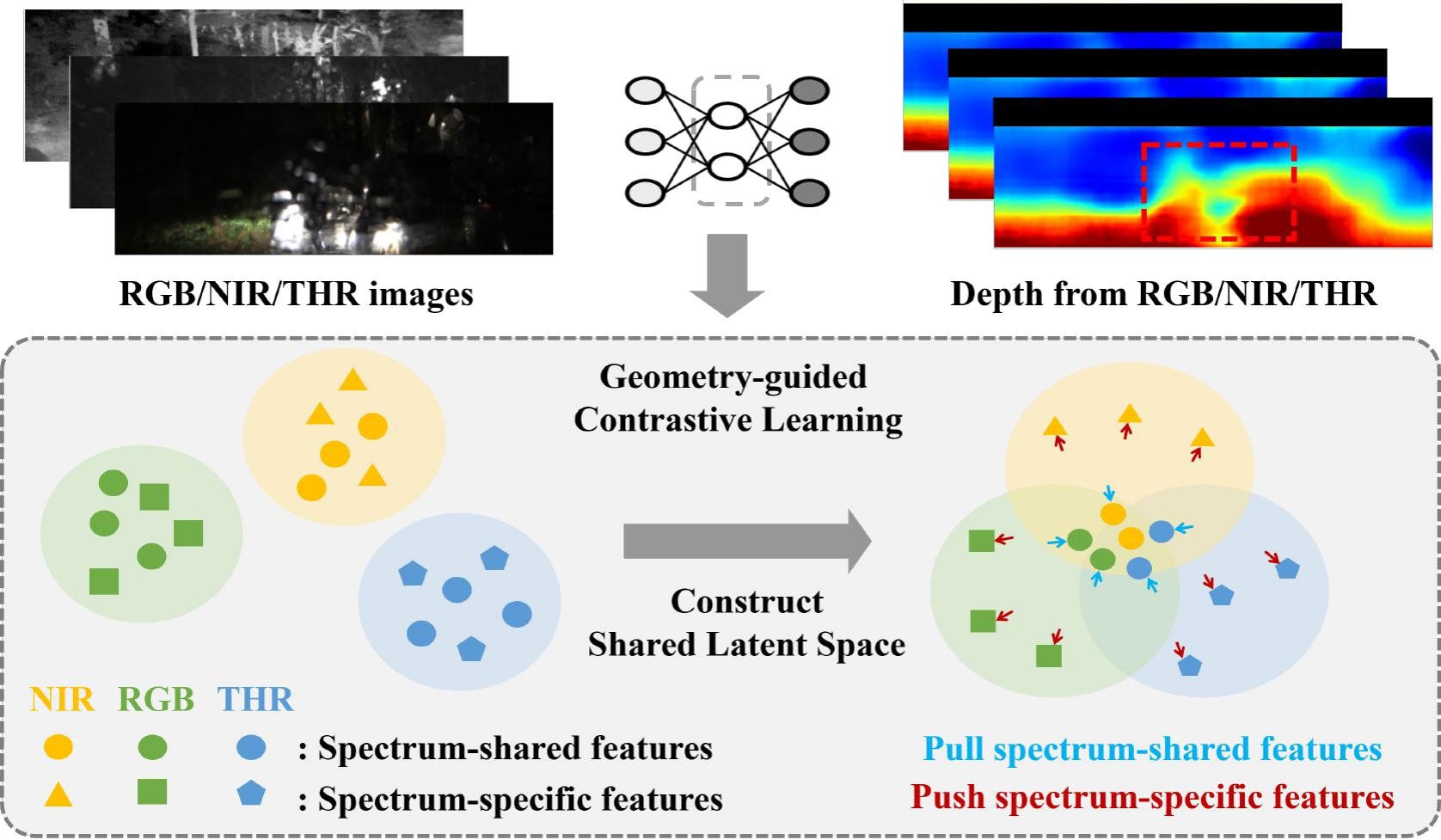}} & 
\multicolumn{2}{c}{\includegraphics[width=0.44\linewidth]{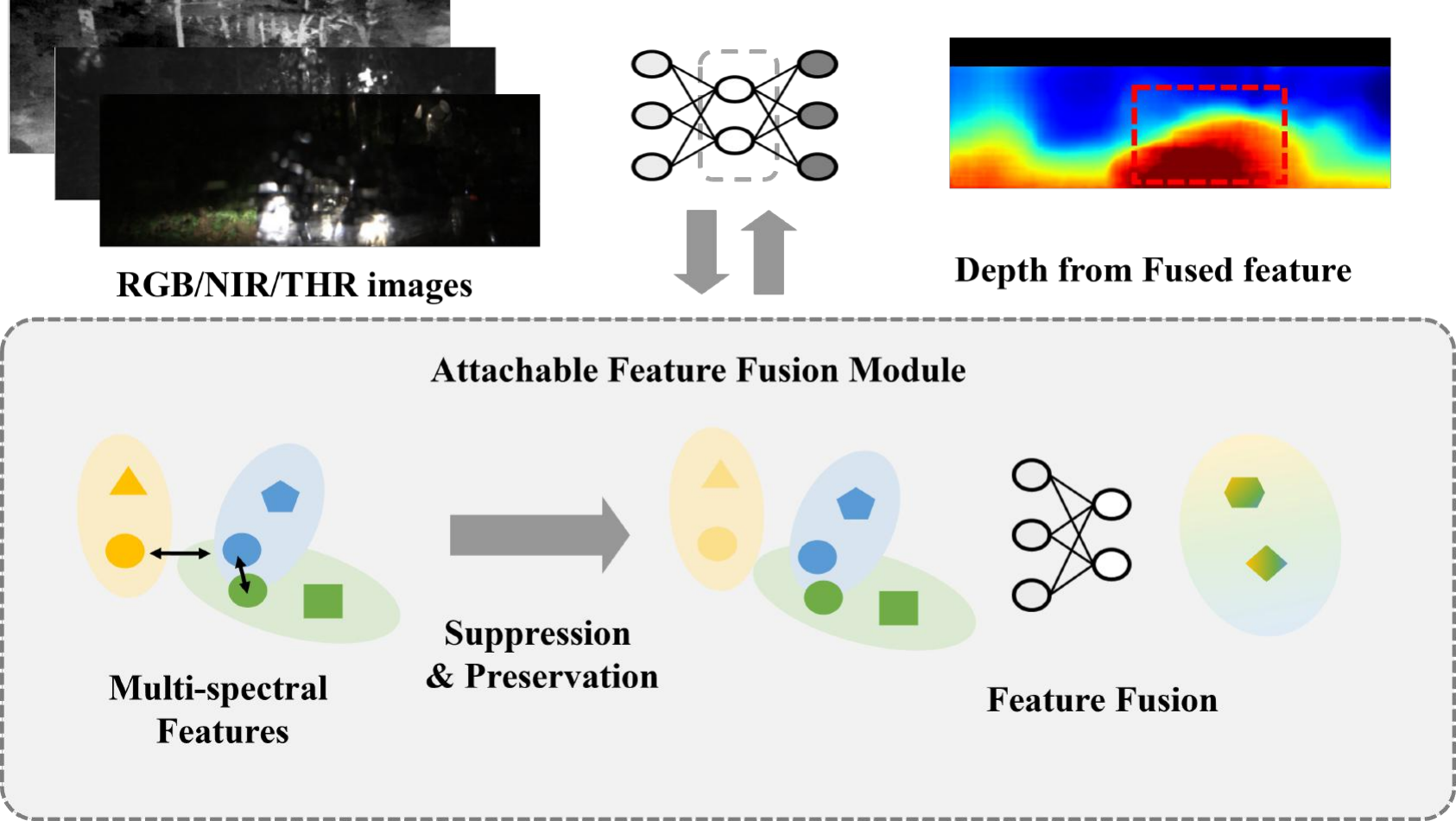}} \\
\multicolumn{2}{c}{{\footnotesize (a) Spectrum generalization via geometry-guided contrastive learning}} & \multicolumn{2}{c}{{\footnotesize (b) Multi-spectral feature fusion via attachable fusion module}} \\
\includegraphics[width=0.22\linewidth]{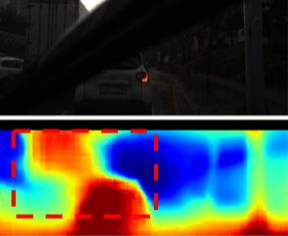} & 
\includegraphics[width=0.22\linewidth]{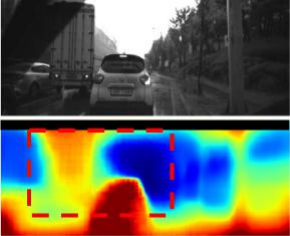} & 
\includegraphics[width=0.22\linewidth]{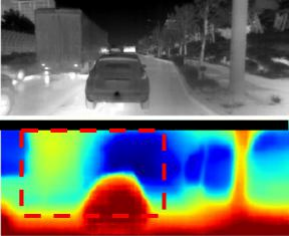} & 
\includegraphics[width=0.22\linewidth]{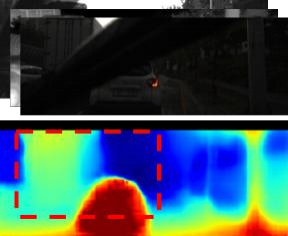} \\
{\footnotesize \textbf{Depth from RGB} } & {\footnotesize \textbf{Depth from NIR}} & {\footnotesize \textbf{Depth from THR} } & {\footnotesize \textbf{Depth from Multi-spectrum}} \\
\multicolumn{4}{c}{{\footnotesize (c) Depth estimation from modality-wise and multi-modality inputs}} \\ 
\end{tabular}
}
\captionof{figure}{{\bf Spectral-wise and multi-spectral fused depth estimation}. 
Our proposed method makes a single network that can estimate spectral-wise depth maps from each different spectral image. 
Also, the proposed attachable fusion module makes the network estimate a reliable and robust depth map under various adverse environments without degeneration of the spectrum generalization ability and modification of the original off-the-shelf network architecture.
}
\end{center}
}]
\thispagestyle{empty}
\pagestyle{empty}

\begin{abstract}
Deploying depth estimation networks in the real world requires high-level robustness against various adverse conditions to ensure safe and reliable autonomy.
For this purpose, many autonomous vehicles employ multi-modal sensor systems, including an RGB camera, NIR camera, thermal camera, LiDAR, or Radar.
They mainly adopt two strategies to use multiple sensors: modality-wise and multi-modal fused inference.
The former method is flexible but memory-inefficient, unreliable, and vulnerable.
Multi-modal fusion can provide high-level reliability, yet it needs a specialized architecture.
In this paper, we propose an effective solution, named \textit{align}-and-\textit{fuse} strategy, for the depth estimation from multi-spectral images. 
In the \textit{align} stage, we align embedding spaces between multiple spectrum bands to learn shareable representation across multi-spectral images by minimizing contrastive loss of global and spatially aligned local features with geometry cue. 
After that, in the \textit{fuse} stage, we train an attachable feature fusion module that can selectively aggregate the multi-spectral features for reliable and robust prediction results.
Based on the proposed method, a single-depth network can achieve both spectral-invariant and multi-spectral fused depth estimation while preserving reliability, memory efficiency, and flexibility. 
\end{abstract}

\section{Introduction}
\label{sec:intro}

Reliable and precise spatial understanding is the most fundamental prior condition for level 5 autonomous driving against adverse weather and lighting conditions, such as rain, fog, dust, haze, and low-light environments. 
Therefore, numerous autonomous vehicle platforms~\cite{huang2022multi,zhang2021autonomous,barnes2020oxford, choi2018kaist,caesar2020nuscenes} concurrently perform perception tasks from multi-modal inputs (\eg, RGB camera, thermal camera, LiDAR, and Radar).
Usually, they employ two main strategies to exploit multiple modality sensors: modality-wise and modality-fused inference~\cite{huang2022multi, zhang2021autonomous, danaci2022survey}.
The former conducts a target task independently with each modality input. 
Therefore, it has high-level flexibility that doesn't tie in with other modality inputs, yet robustness and reliability are not ensured.
The latter can achieve high-level robustness, reliability, and performance.
However, it suffers from an alignment problem between sensors, the need for fusion-oriented architecture, and low flexibility tied with other modality inputs.

Furthermore, modern perception tasks mostly deploy a deep neural network and data-driven machine learning technique to achieve high-level accuracy.
The two problems arise between multi-model sensors: 1) the domain gap between each modality and 2) the absence of general representation.
Usually, most neural networks are designed for each modality and don't consider generalization ability across other heterogeneous modalities.
Also, some specific modality (\eg, RGB image) gets high-level performance boosting based on a large-scale dataset pre-trained backbone network by leveraging learned general representation.
On the other hand, non-conventional sensors (\eg, NIR and thermal images) cannot leverage the general feature representation of a pre-trained backbone network due to the absence of large-scale datasets.

Therefore, we need effective solutions for domain generalization across multi-modal sensors and un-constrained multi-sensor fusion while achieving high-level flexibility, performance, robustness, and reliability.
To this end, in this paper, we propose a multi-spectral generalization method for the monocular depth estimation task and an attachable fusion module that can boost performance while not degrading the learned generalization ability.
Our proposed method can be applied to common monocular depth estimation networks, and the trained network can infer a depth map from both single-modal input and multi-modal input.
Our contributions are summarized as follows:

\begin{itemize}
\item 
We propose a multi-spectral generalization method that learns a shared representation across multi-spectral feature spaces by minimizing contrastive losses of multi-spectral global features and spatially aligned local features for spectral-invariant monocular depth estimation.
\item 
We propose an attachable fusion module that can adaptively aggregate multi-spectral features by preserving reliable features and suppressing suspicious features based on a spectral-shared feature consistency.
\item
We validate that our proposed method can be applied to common monocular depth networks (\eg, MiDaS and NeWCRF) and shows high-level robustness and performance in various sensors and test environments.
\end{itemize}

\section{Related Works}
\label{sec:related works}

\subsection{Deep Monocular Depth Estimation}
{\bf Visible Spectral Band.} 
Most standard network architectures for Monocular Depth Estimation (MDE) have been proposed for the visible spectral band.
The taxonomy of MDE networks is roughly categorized into the methods utilizing per-pixel regression~\cite{lee2019big,Ranftl2022, ranftl2021vision, yuan2022neural}, per-pixel classification~\cite{diaz2019soft, fu2018deep}, and classification-and-regression~\cite{bhat2021adabins, li2022binsformer} head. 
The networks utilize convolution layer~\cite{fu2018deep,lee2019big,Ranftl2022} and transformer block~\cite{bhat2021adabins,Ranftl2021,li2022binsformer,yuan2022neural} to project input image into non-linear feature space. 
After that, the projected features are mapped to a depth map via their prediction head (\eg, regression, classification, or combined heads).

{\bf Thermal Spectral Band.}
The thermal spectral band can provide high-level robustness against various adverse weather and lighting conditions, such as rain, fog, dust, haze, and low-light conditions. 
However, different from the visible spectral band, the deep MDE task for the thermal image has been relatively less studied. 
Recently, a few works~\cite{kim2018multispectral,lu2021alternative,shin2021self,shin2022maximize,shin2023self,shin2023joint} have been proposed in the self-supervised MDE field. 
They mainly utilize extra modality information to train the MDE network for thermal images, such as aligned RGB images~\cite{kim2018multispectral,shin2021self}, image translation network~\cite{lu2021alternative}, and adversarial learning~\cite{shin2023self,shin2023joint} between RGB and thermal images. 
On the other hand, the research~\cite{shin2022maximize} trains an MDE network only with a thermal video.

\subsection{Depth Estimation from Multi-sensor Fusion}
Multi-sensor fusion is a fundamental way to achieve robustness and higher performance for various computer vision and robotics applications.
The most widely used sensors for depth estimation are a RGB camera, LiDAR, and Radar.
Therefore, numerous RGB-Lidar and RGB-Radar fusion researches for depth estimation~\cite{long2021radar,guizilini2021sparse,zhang2018deep,xu2019depth,tang2020learning,park2020non,imran2019depth,gasperini2021r4dyn} have been proposed, known as the depth completion task.
Also, a few RGB-NIR and thermal-LiDAR fusion methods~\cite{zhi2018deep,park2022adaptive,shin2019sparse,kim2024exploiting} have been proposed for a robust solution against low-light and adverse weather conditions.
However, these lines of solutions are designed for sensor fusion purposes, so they need a specialized network architecture.
 
\section{Proposed Method: Align-and-Fuse Strategy} 
\label{sec:method}
\subsection{Method Overview}
\begin{figure*}[t]
\begin{center}
{
\begin{tabular}{c}
\includegraphics[width=0.9\linewidth]{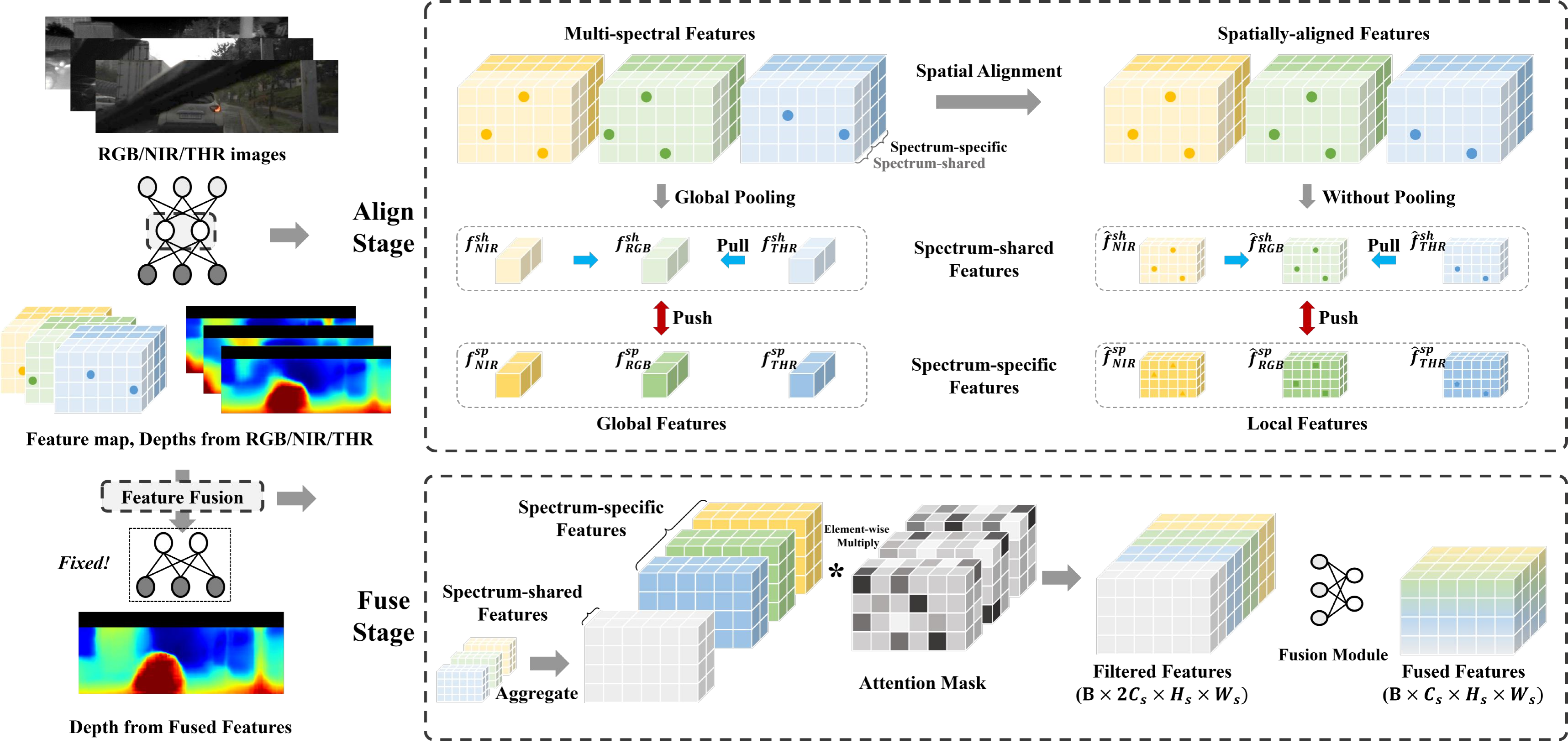} \\
\end{tabular}
}
\end{center}
\vspace{-0.1in}
\caption{{\bf Overall pipeline of our proposed training framework}. 
Our proposed method trains an MDE network in a two-stage learning strategy, named \textit{align}-and-\textit{fuse}.
In the \textit{align} stage, the MDE network learns shared feature representation via geometry-guided contrastive learning by aligning latent spaces of multi-spectral images.
After the first stage training, the \textit{fuse} stage trains an attachable feature fusion module that can selectively aggregate the multi-spectral features for reliable and robust prediction results.
}
\label{fig:overall_pipeline}
\vspace{-0.1in}
\end{figure*}

Our proposed method trains and extends an off-the-shelf network to infer a depth map from both single and multi-modal input (\ie, RGB, NIR, and thermal image). 
To this end, we proposed a two-stage training method, named "\textit{Align}-and-\textit{Fuse}," as shown in~\cref{fig:overall_pipeline}.

{\bf Align Stage} aims to learn general representation across multi-spectrum domains for spectral-invariant depth estimation.
More specifically, we divide each latent feature of multi-spectral images into spectral-shared and spectral-specific features.
After that, we minimize contrastive losses of global and spatially aligned local features of multi-spectral images to make spectral-shared and spectral-specific feature spaces.
The training objective for this stage is as follows: 
\begin{equation}
    L_{align} = L_{sup} + \lambda_{cont}((1-\gamma)L_{g} + \gamma L_{l}),
\end{equation}
where $L_{sup}$ is a supervised loss of a target MDE model, $L_{global}$ is a contrastive loss between the global features of each modality, and $L_{local}$ is a contrastive loss between the spatially-aligned local features from each modality. 

{\bf Fuse Stage} aims to estimate a reliable and robust depth map from multi-spectral images without degenerating generalization performance and modifying off-the-shelf model architecture.
Given feature maps of multi-spectral images, the fusion module suppresses suspicious features via a spectral-shared feature consistency mask.
After that, the weighted feature maps are aggregated via a transformer-based attachable fusion module.
The training objective for the fusion module is as follows: 
\begin{equation}
    L_{fusion} = L_{sup} + \lambda_{geo}L_{geo},
\end{equation}
where $L_{sup}$ is a supervised learning loss of predicted depth to GT depth and $L_{geo}$ is a geometric consistency loss among the modalities. 
Please note that, in this stage, the weights of the monocular depth network are frozen and don't update.

\subsection{Align Stage: Multi-spectrum Generalization}
\subsubsection{Spatial Feature Alignment with Geometric Cues}
Given multi-spectrum images ${I_{rgb},I_{nir},I_{thr}}$, we extract each modality feature ${f_{rgb},f_{nir},f_{thr}}$ and depth map ${D_{rgb},D_{nir},D_{thr}}$ by feeding them to a monocular depth network. 
Since the multi-spectrum images are not spatially aligned because of different field-of-view, sensor size, and spatial resolution, we explicitly align each feature map to share the same spatial coordinate system.
The alignment process utilizes an inverse warping method~\cite{jaderberg2015spatial}.
\begin{equation}
F_{tgt\rightarrow ref}(p_i) = K_{ref}T_{tgt}^{ref}D_{tgt}K^{-1}_{tgt}p_{tgt, i},
\label{equ:inv_warping_a}
\end{equation}
\begin{equation}
\hat{f}_{ref} = W_{inv}(f_{ref}, F_{tgt\rightarrow ref}),   
\label{equ:inv_warping_b}
\end{equation}
where $F_{tgt\rightarrow ref}$ is projection flow from target to reference image plane, $K$ is intrinsic matrix, $T$ is extrinsic matrix, $D$ is estimated depth map of target image plane, $W_{inv}$ is differentiable inverse warping function~\cite{jaderberg2015spatial}, and $\hat{f}_{ref}$ is spatially aligned feature map. 
Here, we align each feature map in the thermal image plane because the thermal image has the largest field of view. Therefore, all spatial features can be projected without information loss.

\subsubsection{Latent Alignment via Contrastive Learning}
For both multi-spectral generalization and effective feature fusion, we design the front half-channel of each feature map to embed spectral-shared features and the remaining half-channel to embed spectral-specific features.
To this end, we exploit the idea of contrastive learning of both global and dense local features.

{\bf Global Contrastive Loss.} 
Each image's global feature is estimated by using a global average polling operation in the last layer of the backbone network.
We set the shared RGB feature $f^{sh}_{rgb}$, as a query $q$ (\ie, anchor point).
Our insight is that most depth estimation networks utilize a pre-trained backbone with large-scale RGB datasets (\eg, ImageNet~\cite{deng2009imagenet}).
Therefore, the RGB feature distribution is well established and has a chance to distill the pre-trained general representation to other non-conventional sensors.

We set global features of different spectrum's shared features $f^{sh}_{nir}, f^{sh}_{thr}$ as the positive keys $k_+$.
For the negative keys $k_-$, we utilize global features of each modality-specific feature $f^{sp}_{rgb}, f^{sp}_{nir}, f^{sp}_{thr}$.
We employ a contrastive loss function InfoNCE~\cite{oord2018representation} to pull query $q$ close to positive keys while pushing it away from other negative keys:
\begin{equation}
L_{g} = -\log\frac{\sum_{k_+}\text{exp}(q\cdot k_+/\tau)}{\sum_{k_+}\text{exp}(q\cdot k_+) +\sum_{k_-}\text{exp}(q\cdot k_-/\tau)}, 
\end{equation}
where $\tau$ denotes a temperature term. 
The query key is detached from the gradient graph to serve as an anchor point.

{\bf Dense Local Contrastive Loss.} 
As shown in~\cref{fig:overall_pipeline}, after the spatial alignment process, we explicitly know the dense matching relation between aligned multi-spectral features ${\hat{f}_{rgb}, \hat{f}_{nir}, \hat{f}_{thr}}$. 
We define each $s$-th feature from the spatially-aligned RGB features ${\hat{f}^{sh}_{rgb}}$ as a query $r^s$. 
Here, all queries belong to the set $S$, where $S$ represents the collection of all per-pixel features in ${\hat{f}^{sh}_{rgb}}$.
The total number of feature vectors in $S$ is given by $|S|=S_h \times S_w$, where $S_h$ and $S_w$ denote the spatial size of the aligned feature map.
Similar to global contrastive loss, we assign spectral-specific features as negative keys $t_-$ and spectral-shared features as positive keys $t_+$.
The dense contrastive loss~\cite{wang2021dense} is defined as:
\begin{equation}
L_{l} = \frac{1}{|S|}\sum_{S}-\log\frac{\sum_{k_+}\text{exp}(r^s\cdot t_+^s/\tau)}{\sum_{k_+}\text{exp}(r^s\cdot t_+^s) +\sum_{k_-^s}\text{exp}(r^s\cdot k_-^s/\tau)},
\end{equation}
where $\tau$ denotes a temperature term. 
Via the \textit{align} stage, the network can have spectral-shared and specific embedding space that is effectively utilized to \textit{fuse} stage.

\subsection{Fuse Stage: Selective Multi-spectrum Fusion}
\subsubsection{Attention Mask for Reliable Feature Selection}
After completing \textit{align} stage training, we conduct \textit{fusion} stage to train the multi-spectral feature fusion module. 
Each modality has advantages and weaknesses depending on the characteristics of the spectrum band. 
Therefore, the reliability and robustness may vary for given lighting and weather environments.
For example, RGB images can provide high-frequency information such as texture, color, and sharp structure details, yet this information is unreliable in low-light conditions.
Therefore, it is necessary to suppress the unreliable feature and preserve the trustful feature.

For this purpose, we exploit cross-spectral shared embedding space that is formed in \textit{align} stage.
After the alignment, each spectrum's feature should be closely located in the shared embedding space.
Also, we utilize one prior knowledge that the thermal image generally has strong reliability and robustness against various adverse weather conditions~\cite{shin2023deep}.
Based on this knowledge, we estimate attention mask for the reliable feature selection, as follows:
\begin{align}
    M_{tgt} = \frac{\hat{f}_{thr}^{sh}\cdot \hat{f}_{tgt}^{sh}}{\text{max}(||\hat{f}_{thr}^{sh}||_2\cdot ||\hat{f}_{tgt}^{sh}||_2, \epsilon)},
\end{align}
where $tgt$ denotes one of spectrum from the spectrum group $\{rgb,nir,thr\}$ and $\epsilon$ is a non-zero value to prevent zero-division. 
If the value of the attention mask closes to 0, it means two shared features of the same scene are totally different, and it could be a potentially unreliable region.

\subsubsection{Attention-weighted Feature Aggregation}
After estimating the attention mask for each modality, we aggregate the attention-weighted features via the feature fusion module.
The feature fusion module consists of one Swin transformer block~\cite{liu2021swin} and an MLP layer to project the input feature $(2C)$ into the original channel dimension $(C)$. 
\begin{gather}
    f_{cat} = [\hat{f}_{fuse}^{sh}, M_{rgb}\cdot \hat{f}_{rgb}^{sp},  M_{nir}\cdot \hat{f}_{nir}^{sp}, M_{thr}\cdot\hat{f}_{thr}^{sp}], \\
    f_{fused} = \theta_{fuse}(f_{cat}),
\end{gather}
where $f_{cat}$ is concatenated feature of spectral-shared and specific features, $\hat{f}_{fuse}^{sh}$ is defined as $\sum_{tgt} M_{tgt}\cdot \hat{f}_{tgt}^{sh}$, and $\theta_{fuse}$ is the weights of feature fusion module.
After that, the fused feature $f_{fused}$ is fed to the decoder to estimate depth map $D_{fuse}$.

\subsubsection{Cross-spectral Geometric Consistency}
Given three spectrum images, the fusion modules can estimate fused features for each spectrum image plane (\ie, $f_{fuse}^{rgb}, f_{fuse}^{nir}, f_{fuse}^{thr}$). 
Also, all modality features are aggregated via the feature fusion layer.
The predicted depth map of each coordinate system should have a consistent 3D structure. 
Therefore, we regularize the estimated depth maps to be consistent across multi-spectrum by utilizing geometric consistency loss~\cite{bian2021unsupervised}.
\begin{equation} 
\label{equ:geometric consistency}
L_{geo} = \frac{1}{|V_p|}\sum\frac{|\Tilde{D}_{fuse}^{tgt}-D_{fuse}^{tgt}|}{\Tilde{D}_{fuse}^{tgt}+D_{fuse}^{tgt}}, 
\end{equation}
where $\Tilde{D}_{fuse}^{tgt}$ is the synthesized depth with the~\cref{equ:inv_warping_a} and ~\cref{equ:inv_warping_b} and $|V_p|$ is the number of validly projected pixels.

\section{Experimental Results}
\label{sec:experiments}
\subsection{Implementation Details}
\label{subsec:impl_detail}
\subsubsection{Multi-Spectral Stereo (MS$^2$) Dataset}
\label{subsec:dataset}
We utilize a Multi-Spectral Stereo (MS$^2$) dataset~\cite{shin2023deep} to train and evaluate our proposed method.
The dataset provides about 162K multi-spectrum data pairs of RGB, NIR, thermal camera, LiDAR, and GPS/IMU.
Also, the dataset includes various location diversity (\eg, campus, city, residential area, and road), time diversity (\eg, daytime and nighttime), and weather diversity (\eg, clear-sky, cloudy, and rainy).
We utilize 26K data pairs for training, 4K pairs for validation, and 2.3K, 2.3K, and 2.5K for evaluation of daytime, nighttime, and rainy conditions.

\begin{table*}[t]
\centering
\caption{\textbf{Quantitative results of monocular depth estimation on the MS$^2$ evaluation dataset}. 
We evaluate our proposed method with representative MDE networks (\ie, MiDaS~\cite{Ranftl2022} and NeWCRF~\cite{yuan2022neural}).
The \textit{align} stage makes the single network encompass multi-spectrum generalization ability that can estimate depth map from each different modality input. 
Also, The \textit{fuse} stage shows overall performance improvement over the single-modality inference results. 
Each modality result is averaged over all day, night, and rainy evaluation sets of the MS$^2$ dataset.
The best performance is highlighted in \textbf{bold}.}
\vspace{-0.1in}
\begin{center}
\resizebox{0.90\linewidth}{!}
{
    \def\arraystretch{0.9}
    \footnotesize
    \begin{tabular}{c|c|c|cccc|ccc} 
    \toprule
    \multirow{2}{*}{Method} & \multicolumn{2}{c|}{Modality} &\multicolumn{4}{c|}{\textbf{Error $\downarrow$}} & \multicolumn{3}{c}{\textbf{Accuracy $\uparrow$}}    \\ \cline{2-10}
    &  Train & Test & AbsRel & SqRel & RMSE & RMSElog &  $\delta < 1.25$ & $\delta < 1.25^{2}$ & $\delta < 1.25^{3}$ \\ 
    \hline
    \multirow{4}{*}{MiDaS-v2.1(s)~\cite{Ranftl2022}}
        & RGB & RGB & 0.122 & 0.858 & 4.533 & 0.162 & 0.857 & 0.971 & 0.992 \\
        & NIR & NIR & 0.129 & 0.818 & 4.178 & 0.170 & 0.846 & 0.969 & 0.990 \\ 
        & THR & THR & 0.100 & 0.428 & 3.312 & 0.132 & 0.901 & 0.984 & 0.996 \\ \cline{2-10}
        &\multicolumn{2}{c|}{\cellcolor{Gray1}Avg} & \cellcolor{Gray1}0.117 & \cellcolor{Gray1}0.701 & \cellcolor{Gray1}4.008 & \cellcolor{Gray1}0.155 & \cellcolor{Gray1}0.868 & \cellcolor{Gray1}0.975 & \cellcolor{Gray1}0.993 \\ \hline
    \multirow{4}{*}{MiDaS + \textit{Align}} & \multirow{3}{*}{Multi}
           & RGB & 0.115 & 0.722 & 4.269 & 0.154 & 0.873 & 0.976 & 0.994 \\
        &  & NIR & 0.122 & 0.791 & 4.099 & 0.157 & 0.864 & 0.973 & 0.992 \\
        &  & THR & 0.088 & 0.387 & 3.161 & 0.120 & 0.922 & 0.989 & 0.998 \\
        &\multicolumn{2}{c|}{\cellcolor{Gray1}Avg} & \cellcolor{Gray1}0.108 &  \cellcolor{Gray1}0.633 &  \cellcolor{Gray1}3.843 &  \cellcolor{Gray1}0.144 &  \cellcolor{Gray1}0.886 &  \cellcolor{Gray1}0.979 &  \cellcolor{Gray1}0.995 \\ \hline
    \multirow{4}{*}{MiDaS + \textit{Fuse}} & \multirow{3}{*}{Multi} & \multirow{3}{*}{Multi}
            & 0.107 & 0.607 & 3.989 & 0.143 & 0.890 & 0.983 & 0.996 \\
        & & & 0.110 & 0.573 & 3.694 & 0.144 & 0.887 & 0.982 & 0.996 \\
        & & & 0.086 & 0.382 & 3.090 & 0.122 & 0.928 & 0.990 & 0.998 \\
        &\multicolumn{2}{c|}{\cellcolor{Gray1}Avg} & \cellcolor{Gray1}\textbf{0.101} & \cellcolor{Gray1}\textbf{0.521} & \cellcolor{Gray1}\textbf{3.591} & \cellcolor{Gray1}\textbf{0.136} & \cellcolor{Gray1}\textbf{0.902} & \cellcolor{Gray1}\textbf{0.985} & \cellcolor{Gray1}\textbf{0.997} \\
    \bottomrule
    \toprule
    \multirow{4}{*}{NeWCRF~\cite{yuan2022neural}}
        & RGB & RGB & 0.099 & 0.520 & 3.729 & 0.133 & 0.905 & 0.987 & 0.997 \\
        & NIR & NIR & 0.112 & 0.641 & 3.791 & 0.144 & 0.883 & 0.979 & 0.994 \\
        & THR & THR & 0.081 & 0.331 & 2.937 & 0.109 & 0.937 & 0.992 & 0.999 \\ \cline{2-10}
        &\multicolumn{2}{c|}{\cellcolor{Gray1}Avg} & \cellcolor{Gray1}0.097 & \cellcolor{Gray1}0.497 & \cellcolor{Gray1}3.486 & \cellcolor{Gray1}0.128 & \cellcolor{Gray1}0.908 & \cellcolor{Gray1}0.986 & \cellcolor{Gray1}0.997 \\ \hline
    \multirow{4}{*}{NeWCRF + \textit{Align}} & \multirow{3}{*}{Multi}
           & RGB & 0.097 & 0.504 & 3.657 & 0.130 & 0.910 & 0.987 & 0.997 \\
        &  & NIR & 0.107 & 0.571 & 3.617 & 0.139 & 0.896 & 0.983 & 0.995 \\
        &  & THR & 0.079 & 0.310 & 2.860 & 0.108 & 0.940 & 0.994 & 0.999 \\ \cline{2-10}
        &\multicolumn{2}{c|}{\cellcolor{Gray1}Avg} & \cellcolor{Gray1}0.094 & \cellcolor{Gray1}0.462 & \cellcolor{Gray1}3.378 & \cellcolor{Gray1}0.126 & \cellcolor{Gray1}0.915 & \cellcolor{Gray1}0.988 & \cellcolor{Gray1}0.997 \\ \hline
    \multirow{4}{*}{NeWCRF + \textit{Fuse}} & \multirow{3}{*}{Multi} & \multirow{3}{*}{Multi}
             & 0.087 & 0.408 & 3.366 & 0.119 & 0.928 & 0.992 & 0.998 \\
        &  & & 0.095 & 0.423 & 3.255 & 0.125 & 0.917 & 0.990 & 0.997 \\
        &  & & 0.072 & 0.251 & 2.623 & 0.098 & 0.954 & 0.996 & 1.000 \\
        &\multicolumn{2}{c|}{\cellcolor{Gray1}Avg} & \cellcolor{Gray1}\textbf{0.085} & \cellcolor{Gray1}\textbf{0.361} & \cellcolor{Gray1}\textbf{3.081} & \cellcolor{Gray1}\textbf{0.114} & \cellcolor{Gray1}\textbf{0.933} & \cellcolor{Gray1}\textbf{0.993} & \cellcolor{Gray1}\textbf{0.998} \\
    \bottomrule
    \end{tabular} 
}
\vspace{-0.2in}
\end{center}
\label{tab:quantitative_result}
\end{table*}

\subsubsection{Network Architecture} 

{\bf Monocular Depth Estimation (MDE) Network.} 
We adopted MiDaS-v2.1~\cite{Ranftl2022} and NeWCRF~\cite{yuan2022neural} to evaluate our proposed method.
We utilize each off-the-shelf network architecture from their official source code and don't modify any architecture details.
For the NIR and thermal images that have only a single channel information, we repeat them three times along the channel axis to be identical to the RGB image.
All MDE networks are initialized with ImageNet pre-trained backbone model~\cite{deng2009imagenet} by following their original implementations~\cite{Ranftl2022,yuan2022neural}.

{\bf Feature Fusion Module.} 
The module consists of a single Swin transformer block~\cite{liu2021swin} with four multi-head attention and a single MLP layer.
The transformer layer effectively aggregates multi-spectral features via a multi-head attention mechanism. 
After that, the MLP layer projects an aggregated feature $(B\times 2C \times H_s \times W_s)$ into the original input channel dimension $(B\times C \times H_s \times W_s)$ for the MDE decoder head, where $s$ is a spatial scale factor.

\subsubsection{Training Details}
We utilize the PyTorch library to implement our proposed method.
All models are trained for 60 epochs on a single RTX Titan GPU with 24GB memory.
We utilize a batch size of 8 for the single modality training of all MDE models and a batch size of 4 for multi-modality training, including the Baseline model, \textit{align} stage, and \textit{fusion} stage training.
During \textit{align} stage training, MDE networks are trained with each model's supervised loss~\cite{Ranftl2022, yuan2022neural} and the proposed contrastive learning scheme.
The \textit{fuse} stage only updates the weight of feature fusion modules so that it does not degenerate the learned generalization ability of the MDE network by freezing MDE network weights.

We utilize AdamW optimizer~\cite{loshchilov2018decoupled} with an initial learning rate of $1e^{-4}$ for all model training.
For the data augmentation, we apply random center crop-and-resize, brightness jitter, and contrast jitter for all modalities.
Saturation and hue jitters~\cite{liu2016ssd} are additionally applied to the RGB modality.
Also, the horizontal flip is applied to the single-modality training.
The scale factors ($\lambda_{cont}$ and $\lambda_{geo}$) are set to $0.01$ and $0.5$. 
The balance factor $\gamma$ is set to $0.5$, the same as the DenseCL~\cite{wang2021dense}.

\subsection{Monocular Depth Estimation from Multiple Spectrum}
\label{subsec:exp_result}

{\bf Spectral-wise Depth Estimation.} 
We compare the modality-wise networks and our modality-generalized network (\ie, \textit{+align}).
Generally, a na\"ive joint learning of multiple modalities tends to degenerate the network performance because of domain conflict, as shown in experimented in~\cref{subsec:ablation_study}.
However, our proposed method (\ie, +\textit{align}) makes the MDE networks learn the generalizable representation via contrastive learning between shared and specific features of each multi-spectral domain.
Also, the contrastive loss in the spectrum shared space leads to a performance improvement by providing a cross-supervision between the multi-spectral modalities.

\begin{figure*}[t]
\begin{center}
{
\begin{tabular}{c@{\hskip 0.01\linewidth}c}
\includegraphics[width=0.49\linewidth]{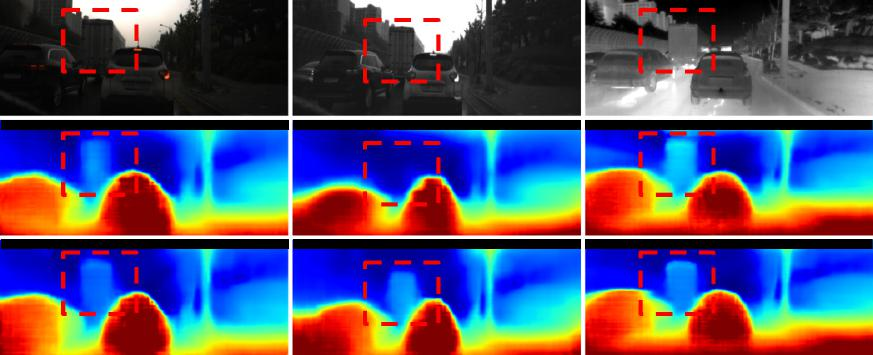} &
\includegraphics[width=0.49\linewidth]{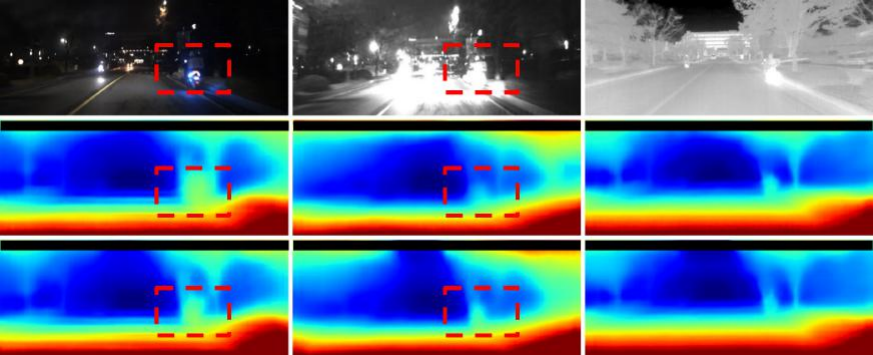} \\ 
{\footnotesize (a) Day (1): spectral-wise ($2^{nd}$) and fused ($3^{rd}$) depth estimation } & {\footnotesize (b) Night (1): spectral-wise ($2^{nd}$) and fused ($3^{rd}$) depth estimation } \\
\includegraphics[width=0.49\linewidth]{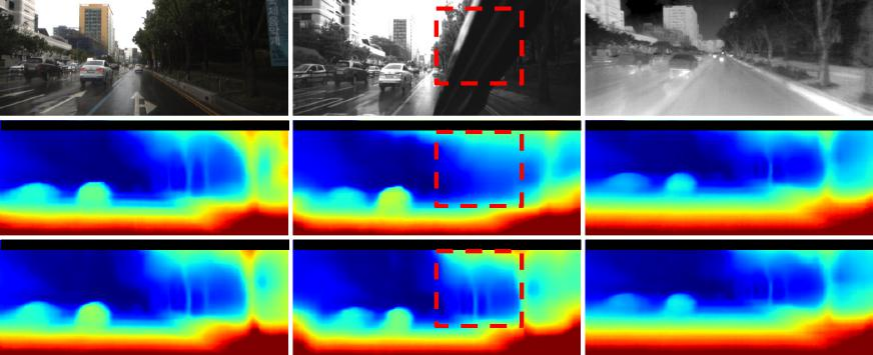} &
\includegraphics[width=0.49\linewidth]{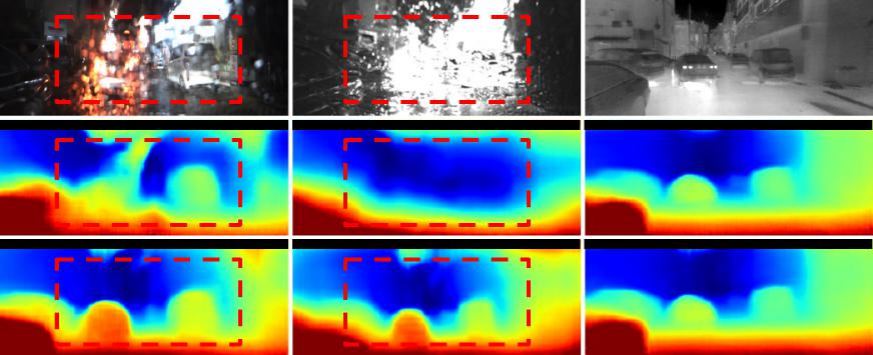} \\ 
{\footnotesize (e) Rain (1): spectral-wise ($2^{nd}$) and fused ($3^{rd}$) depth estimation } & {\footnotesize (f) Rain (2): spectral-wise ($2^{nd}$) and fused ($3^{rd}$) depth estimation } \\
\end{tabular}
}
\end{center}
\vspace{-0.10in}
\caption{{\bf Qualitative comparison of spectral-wise and multi-spectral depth estimation in day, night, and rainy conditions}. 
Our proposed method enables a single network to estimate depth maps from each spectrum input. 
Also, the fusion modules make the network achieve robust and reliable depth estimation results against rain, occlusion, and glare effects.
}
\label{fig:qualitative_results}
\vspace{-0.1in}
\end{figure*}

{\bf Multi-Spectral Fused Depth Estimation.}
As shown in~\cref{fig:qualitative_results} and ~\cref{tab:quantitative_result}, the proposed feature fusion module (\ie, +\textit{Fuse}) brings substantial performance boosting for all MDE models by effectively aggregating multi-spectral features in a reliable way.
Furthermore, in low-light and rainy conditions, the MDE networks can estimate a robust depth map via the help of the feature fusion module (\cref{fig:qualitative_results}).
Also, as shown in~\cref{tab:ablation2} and~\cref{fig:ablation_study}, the multi-spectral feature fusion shows high effectiveness for the hostile condition of each modality.
Specifically, RGB spectrum suffers from blurry and low-contrast image details because of insufficient lighting conditions at the night evaluation set.
However, when the other spectrum features are available, the performance is greatly improved by complementing unreliable feature regions of the RGB spectrum with other reliable spectrum features.
Similarly, in rainy conditions, the fusion modules make the network achieve robust and reliable depth estimation results against rain, occlusion, and glare effects.
As a result, the proposed methods (\ie, \textit{Align}-and-\textit{Fuse}) enable an arbitrary MDE network to estimate the depth map from each spectrum input and multi-spectrum input.

\begin{table}[t]
\centering
\caption{\textbf{Ablation study of the proposed method}. 
} \vspace{-0.1in}
\begin{center}
\resizebox{0.98\linewidth}{!}
{
    \def\arraystretch{1.2}
    \footnotesize
    \begin{tabular}{c|c|cc|c|cc} 
    \toprule
    \multicolumn{2}{c|}{\multirow{2}{*}{Method}} & \multicolumn{2}{c|}{Modality} & \multirow{2}{*}{RMSE ($\downarrow)$} & \multicolumn{2}{c}{\textbf{Accuracy $\uparrow$}} \\ \cline{3-4} \cline{6-7}
    \multicolumn{2}{c|}{} & Train & Test & & $\delta < 1.25$ & $\delta<1.25^3$ \\
    \hline 
    \multicolumn{2}{c|}{Single-Modality} & Single & Single & 4.008 & 0.868 & 0.975 \\ \hline
    \multicolumn{2}{c|}{S2R Depth~\cite{chen2021s2r}} & Multi & Single & 3.897 & 0.873 & 0.972 \\ \hline       
    \multicolumn{2}{c|}{Baseline + \textit{fuse}}  & Multi & Multi & 3.823 & 0.888 & 0.983 \\ \hline 
    \multirow{4}{*}{\rotatebox{90}{\textit{align}}} &
       Baseline   & Multi & Single & 3.983 & 0.871 & 0.972 \\
    &  + $L_{g}$  & Multi & Single & 3.914 & 0.881 & 0.978 \\
    & + $L_{l}$   & Multi & Single & 3.894 & 0.884 & 0.977 \\
    & + $L_{g}$ + $L_{l}$  & Multi & Single & \textbf{3.843} & \textbf{0.886} & \textbf{0.979}  \\
    \midrule
    \multirow{2}{*}{\rotatebox{90}{\textit{fuse}}} &
    + Fusion Module & Multi & Multi  & 3.659 & 0.898 & 0.984  \\ 
    & + $L_{geo}$   & Multi & Multi  & \textbf{3.591} & \textbf{0.902} & \textbf{0.985}  \\ 
    \bottomrule
    \end{tabular} 
}
\end{center}
\label{tab:ablation}
\vspace{-0.2in}
\end{table}

\subsection{Ablation Study}
\label{subsec:ablation_study}
{\bf Multi-spectrum Generalization.}
We conduct an ablation study about the multi-spectrum generalization of our proposed method, as shown in~\cref{tab:ablation}.
We utilize MiDaS-v2.1 network~\cite{Ranftl2022} for all ablation studies.
For the Baseline model, we trained the network with na\"ive joint learning of multi-spectral images for depth estimation tasks without any contrastive losses.
We also compared our model with the S2R Depth~\cite{chen2021s2r}. 
Adding global and local contrastive loss stand-alone (\ie, $+L_{g}$ and $+L_{l}$) leads to an overall performance improvement.
Differ from the original dense contrastive loss~\cite{wang2021dense}, we explicitly find the dense relation via spatial alignment process. 
Therefore, without the presence of global contrastive loss $+L_{g}$, the local contrastive loss can provide a performance improvement.
Combining them (\ie, $L_{g}$+$L_{l}$) also further increase the performance. 

\begin{figure}[t]
\begin{center}
{
\begin{tabular}{c@{\hskip 0.01\linewidth}c}
\includegraphics[width=0.48\linewidth]{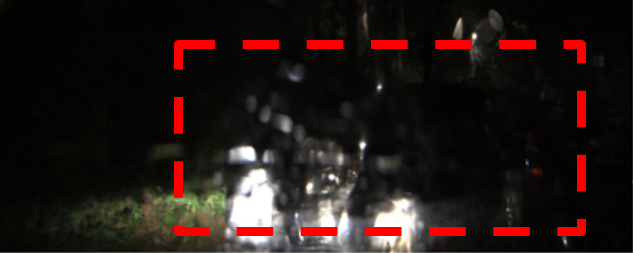} & 
\includegraphics[width=0.48\linewidth]{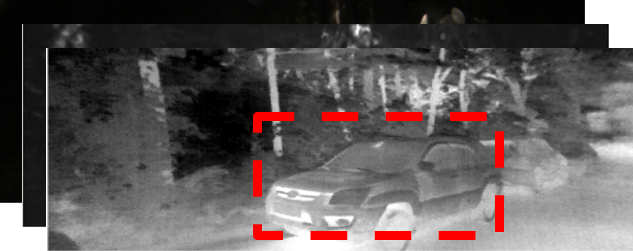} \\
{\footnotesize (a) Single RGB input image} & {\footnotesize (b) Multi-spectrum image}\\
\includegraphics[width=0.48\linewidth]{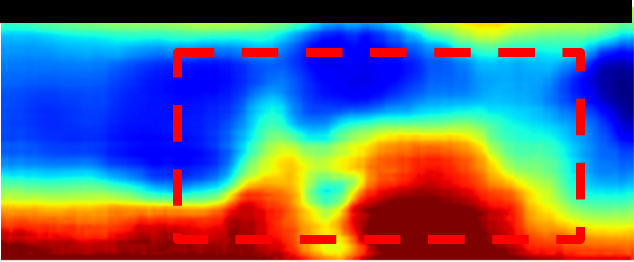} & 
\includegraphics[width=0.48\linewidth]{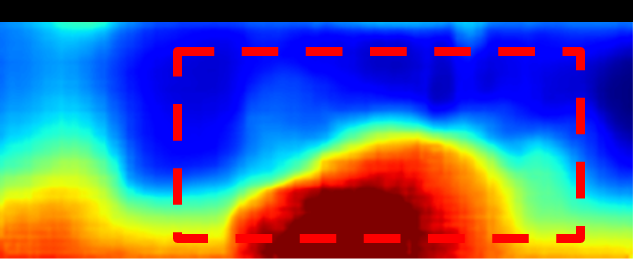} \\
{\footnotesize (c) Depth from RGB image} & {\footnotesize (d) Depth from multi-spectrum }\\
\end{tabular}
}
\end{center}
\vspace{-0.1in}
\caption{{\bf Qualitative comparison with and without the feature fusion module.} 
}
\label{fig:ablation_study}
\vspace{-0.2in}
\end{figure}

{\bf Multi-spectral Feature Fusion.}
The fusion module obviously leads to performance boosting by complementing the weakness of each sensor modality.
Also, enforcing prediction consistency across each sensor coordinate system (\ie, $+L_{geo}$ gets improved performance by supplying extra self-supervision between each other. 
The advantage of multi-spectrum fusion is also observable in~\cref{fig:ablation_study}.
Under low-light conditions, single-modality prediction provides unreliable and inaccurate results.
However, multi-modality prediction has reliable, robust, and accurate depth estimation results.

\begin{table}[t]
\centering
\caption{\textbf{The effect of multi-spectral feature fusion for various environments.}  
} \vspace{-0.1in}
\begin{center}
\resizebox{0.98\linewidth}{!}
{
    \def\arraystretch{1.2}
    \footnotesize
    \begin{tabular}{c|c|c|c|cc} 
    \toprule
    \multirow{2}{*}{Method} & \multirow{2}{*}{Modal} & \multirow{2}{*}{TestSet} & \multirow{2}{*}{RMSE ($\downarrow)$} & \multicolumn{2}{c}{\textbf{Accuracy $\uparrow$}} \\ \cline{5-6}
     & & & & $\delta < 1.25$ & $\delta<1.25^2$ \\
    \hline 
        &  \multirow{4}{*}{\rotatebox{90}{RGB}} 
                          & Day   & 3.013 & 0.952 & 0.994 \\
        NeWCRF          & & Night & 3.458 & 0.907 & 0.991 \\ 
        +\textit{align} & & Rain  & 4.440 & 0.875 & 0.977 \\  \cline{3-6}
        & & \cellcolor{Gray1}Avg  & \cellcolor{Gray1}3.657 & \cellcolor{Gray1}0.911 & \cellcolor{Gray1}0.987 \\
    \midrule 
        & \multirow{4}{*}{\rotatebox{90}{RGB}} 
                         & Day   & 2.927 (\textcolor{red}{-0.086}) & 0.955 (\textcolor{red}{+0.003}) & 0.995 (\textcolor{red}{+0.001}) \\
        NeWCRF         & & Night & 3.224 (\textcolor{red}{\textbf{-0.234}}) & 0.925 (\textcolor{red}{\textbf{+0.018}}) & 0.994 (\textcolor{red}{\textbf{+0.003}}) \\
        +\textit{fuse}  & & Rain  & 3.906 (\textcolor{red}{\textbf{-0.534}}) & 0.906 (\textcolor{red}{\textbf{+0.031}}) & 0.987 (\textcolor{red}{\textbf{+0.010}}) \\ \cline{3-6}
        & & \cellcolor{Gray1}Avg  & \cellcolor{Gray1}3.366 (\textcolor{red}{-0.291})  & \cellcolor{Gray1}0.928 (\textcolor{red}{+0.017}) & \cellcolor{Gray1}0.992 (\textcolor{red}{+0.005}) \\
    \bottomrule
    \end{tabular} 
}
\end{center}
\label{tab:ablation2}
\vspace{-0.2in}
\end{table}

\section{Conclusion}
In this paper, we aim to solve practical issues of deploying a Monocular Depth Estimation (MDE) network in the wild; Spectral-wise depth estimation and Multi-spectral fused depth estimation.
The core idea is establishing shared latent space between the modalities and exploiting the space for generalization and fusion via geometry-guided contrastive learning.
Based on this idea, we validate that our proposed method makes a single MDE network encompass generalization ability across multi-modal sensors and achieve un-constrained multi-sensor fusion while achieving high-level flexibility, performance, and robustness.
Our source code is available at \url{https://github.com/UkcheolShin/BridgeMultiSpectralDepth}.



\section*{ACKNOWLEDGMENT}
This work was supported by a grant (P0026022) from R\&D Program funded by Ministry of Trade, Industry and Energy of Korean government.

{\small
\bibliographystyle{IEEEtrans}
\bibliography{egbib}

\begin{thebibliography}{10}
\providecommand{\url}[1]{#1}
\csname url@rmstyle\endcsname
\providecommand{\newblock}{\relax}
\providecommand{\bibinfo}[2]{#2}
\providecommand\BIBentrySTDinterwordspacing{\spaceskip=0pt\relax}
\providecommand\BIBentryALTinterwordstretchfactor{4}
\providecommand\BIBentryALTinterwordspacing{\spaceskip=\fontdimen2\font plus
\BIBentryALTinterwordstretchfactor\fontdimen3\font minus \fontdimen4\font\relax}
\providecommand\BIBforeignlanguage[2]{{%
\expandafter\ifx\csname l@#1\endcsname\relax
\typeout{** WARNING: IEEEtran.bst: No hyphenation pattern has been}%
\typeout{** loaded for the language `#1'. Using the pattern for}%
\typeout{** the default language instead.}%
\else
\language=\csname l@#1\endcsname
\fi
#2}}

\bibitem{huang2022multi}
K.~Huang, B.~Shi, X.~Li, X.~Li, S.~Huang, and Y.~Li, ``Multi-modal sensor fusion for auto driving perception: A survey,'' \emph{arXiv preprint arXiv:2202.02703}, 2022.

\bibitem{zhang2021autonomous}
Y.~Zhang, A.~Carballo, H.~Yang, and K.~Takeda, ``Autonomous driving in adverse weather conditions: A survey,'' \emph{arXiv preprint arXiv:2112.08936}, 2021.

\bibitem{barnes2020oxford}
D.~Barnes, M.~Gadd, P.~Murcutt, P.~Newman, and I.~Posner, ``The oxford radar robotcar dataset: A radar extension to the oxford robotcar dataset,'' in \emph{2020 IEEE International Conference on Robotics and Automation}.\hskip 1em plus 0.5em minus 0.4em\relax IEEE, 2020, pp. 6433--6438.

\bibitem{choi2018kaist}
Y.~Choi, N.~Kim, S.~Hwang, K.~Park, J.~S. Yoon, K.~An, and I.~S. Kweon, ``Kaist multi-spectral day/night data set for autonomous and assisted driving,'' \emph{IEEE Transactions on Intelligent Transportation Systems}, vol.~19, no.~3, pp. 934--948, 2018.

\bibitem{caesar2020nuscenes}
H.~Caesar, V.~Bankiti, A.~H. Lang, S.~Vora, V.~E. Liong, Q.~Xu, A.~Krishnan, Y.~Pan, G.~Baldan, and O.~Beijbom, ``nuscenes: A multimodal dataset for autonomous driving,'' in \emph{Proceedings of the IEEE/CVF conference on computer vision and pattern recognition}, 2020, pp. 11\,621--11\,631.

\bibitem{danaci2022survey}
K.~I. Danaci and E.~Akagunduz, ``A survey on infrared image and video sets,'' \emph{arXiv preprint arXiv:2203.08581}, 2022.

\bibitem{lee2019big}
J.~H. Lee, M.-K. Han, D.~W. Ko, and I.~H. Suh, ``From big to small: Multi-scale local planar guidance for monocular depth estimation,'' \emph{arXiv preprint arXiv:1907.10326}, 2019.

\bibitem{Ranftl2022}
R.~Ranftl, K.~Lasinger, D.~Hafner, K.~Schindler, and V.~Koltun, ``Towards robust monocular depth estimation: Mixing datasets for zero-shot cross-dataset transfer,'' \emph{IEEE Transactions on Pattern Analysis and Machine Intelligence}, vol.~44, no.~3, 2022.

\bibitem{ranftl2021vision}
R.~Ranftl, A.~Bochkovskiy, and V.~Koltun, ``Vision transformers for dense prediction,'' in \emph{Proceedings of the IEEE/CVF International Conference on Computer Vision}, 2021, pp. 12\,179--12\,188.

\bibitem{yuan2022neural}
W.~Yuan, X.~Gu, Z.~Dai, S.~Zhu, and P.~Tan, ``Neural window fully-connected crfs for monocular depth estimation,'' in \emph{Proceedings of the IEEE/CVF Conference on Computer Vision and Pattern Recognition}, 2022, pp. 3916--3925.

\bibitem{diaz2019soft}
R.~Diaz and A.~Marathe, ``Soft labels for ordinal regression,'' in \emph{Proceedings of the IEEE/CVF conference on computer vision and pattern recognition}, 2019, pp. 4738--4747.

\bibitem{fu2018deep}
H.~Fu, M.~Gong, C.~Wang, K.~Batmanghelich, and D.~Tao, ``Deep ordinal regression network for monocular depth estimation,'' in \emph{Proceedings of the IEEE conference on computer vision and pattern recognition}, 2018, pp. 2002--2011.

\bibitem{bhat2021adabins}
S.~F. Bhat, I.~Alhashim, and P.~Wonka, ``Adabins: Depth estimation using adaptive bins,'' in \emph{Proceedings of the IEEE/CVF Conference on Computer Vision and Pattern Recognition}, 2021, pp. 4009--4018.

\bibitem{li2022binsformer}
Z.~Li, X.~Wang, X.~Liu, and J.~Jiang, ``Binsformer: Revisiting adaptive bins for monocular depth estimation,'' \emph{arXiv preprint arXiv:2204.00987}, 2022.

\bibitem{Ranftl2021}
R.~Ranftl, A.~Bochkovskiy, and V.~Koltun, ``Vision transformers for dense prediction,'' \emph{ICCV}, 2021.

\bibitem{kim2018multispectral}
N.~Kim, Y.~Choi, S.~Hwang, and I.~S. Kweon, ``Multispectral transfer network: Unsupervised depth estimation for all-day vision,'' in \emph{Thirty-Second AAAI Conference on Artificial Intelligence}, 2018.

\bibitem{lu2021alternative}
Y.~Lu and G.~Lu, ``An alternative of lidar in nighttime: Unsupervised depth estimation based on single thermal image,'' in \emph{Proceedings of the IEEE/CVF Winter Conference on Applications of Computer Vision}, 2021, pp. 3833--3843.

\bibitem{shin2021self}
U.~Shin, K.~Lee, S.~Lee, and I.~S. Kweon, ``Self-supervised depth and ego-motion estimation for monocular thermal video using multi-spectral consistency loss,'' \emph{IEEE Robotics and Automation Letters}, vol.~7, no.~2, pp. 1103--1110, 2021.

\bibitem{shin2022maximize}
U.~Shin, K.~Lee, B.-U. Lee, and I.~S. Kweon, ``Maximizing self-supervision from thermal image for effective self-supervised learning of depth and ego-motion,'' \emph{IEEE Robotics and Automation Letters}, vol.~7, no.~3, pp. 7771--7778, 2022.

\bibitem{shin2023self}
U.~Shin, K.~Park, B.-U. Lee, K.~Lee, and I.~S. Kweon, ``Self-supervised monocular depth estimation from thermal images via adversarial multi-spectral adaptation,'' in \emph{Proceedings of the IEEE/CVF Winter Conference on Applications of Computer Vision}, 2023, pp. 5798--5807.

\bibitem{shin2023joint}
U.~Shin, K.~Park, K.~Lee, B.-U. Lee, and I.~S. Kweon, ``Joint self-supervised learning and adversarial adaptation for monocular depth estimation from thermal image,'' \emph{Machine Vision and Applications}, vol.~34, no.~4, p.~55, 2023.

\bibitem{long2021radar}
Y.~Long, D.~Morris, X.~Liu, M.~Castro, P.~Chakravarty, and P.~Narayanan, ``Radar-camera pixel depth association for depth completion,'' in \emph{Proceedings of the IEEE/CVF Conference on Computer Vision and Pattern Recognition}, 2021, pp. 12\,507--12\,516.

\bibitem{guizilini2021sparse}
V.~Guizilini, R.~Ambrus, W.~Burgard, and A.~Gaidon, ``Sparse auxiliary networks for unified monocular depth prediction and completion,'' in \emph{Proceedings of the IEEE/CVF Conference on Computer Vision and Pattern Recognition}, 2021, pp. 11\,078--11\,088.

\bibitem{zhang2018deep}
Y.~Zhang and T.~Funkhouser, ``Deep depth completion of a single rgb-d image,'' in \emph{Proceedings of the IEEE Conference on Computer Vision and Pattern Recognition}, 2018, pp. 175--185.

\bibitem{xu2019depth}
Y.~Xu, X.~Zhu, J.~Shi, G.~Zhang, H.~Bao, and H.~Li, ``Depth completion from sparse lidar data with depth-normal constraints,'' in \emph{Proceedings of the IEEE/CVF International Conference on Computer Vision}, 2019, pp. 2811--2820.

\bibitem{tang2020learning}
J.~Tang, F.-P. Tian, W.~Feng, J.~Li, and P.~Tan, ``Learning guided convolutional network for depth completion,'' \emph{IEEE Transactions on Image Processing}, vol.~30, pp. 1116--1129, 2020.

\bibitem{park2020non}
J.~Park, K.~Joo, Z.~Hu, C.-K. Liu, and I.~So~Kweon, ``Non-local spatial propagation network for depth completion,'' in \emph{European Conference on Computer Vision}.\hskip 1em plus 0.5em minus 0.4em\relax Springer, 2020, pp. 120--136.

\bibitem{imran2019depth}
S.~Imran, Y.~Long, X.~Liu, and D.~Morris, ``Depth coefficients for depth completion,'' in \emph{2019 IEEE/CVF Conference on Computer Vision and Pattern Recognition}.\hskip 1em plus 0.5em minus 0.4em\relax IEEE, 2019, pp. 12\,438--12\,447.

\bibitem{gasperini2021r4dyn}
S.~Gasperini, P.~Koch, V.~Dallabetta, N.~Navab, B.~Busam, and F.~Tombari, ``R4dyn: Exploring radar for self-supervised monocular depth estimation of dynamic scenes,'' in \emph{2021 International Conference on 3D Vision}.\hskip 1em plus 0.5em minus 0.4em\relax IEEE, 2021, pp. 751--760.

\bibitem{zhi2018deep}
T.~Zhi, B.~R. Pires, M.~Hebert, and S.~G. Narasimhan, ``Deep material-aware cross-spectral stereo matching,'' in \emph{Proceedings of the IEEE Conference on Computer Vision and Pattern Recognition}, 2018, pp. 1916--1925.

\bibitem{park2022adaptive}
J.~Park, Y.~Jeong, K.~Joo, D.~Cho, and I.~S. Kweon, ``Adaptive cost volume fusion network for multi-modal depth estimation in changing environments,'' \emph{IEEE Robotics and Automation Letters}, 2022.

\bibitem{shin2019sparse}
Y.-S. Shin and A.~Kim, ``Sparse depth enhanced direct thermal-infrared slam beyond the visible spectrum,'' \emph{IEEE Robotics and Automation Letters}, vol.~4, no.~3, pp. 2918--2925, 2019.

\bibitem{kim2024exploiting}
J.~Kim, U.~Shin, S.~Heo, and J.~Park, ``Exploiting cross-modal cost volume for multi-sensor depth estimation,'' in \emph{Proceedings of the Asian Conference on Computer Vision}, 2024, pp. 1420--1436.

\bibitem{jaderberg2015spatial}
M.~Jaderberg, K.~Simonyan, A.~Zisserman, \emph{et~al.}, ``Spatial transformer networks,'' \emph{Advances in neural information processing systems}, vol.~28, 2015.

\bibitem{deng2009imagenet}
J.~Deng, W.~Dong, R.~Socher, L.-J. Li, K.~Li, and L.~Fei-Fei, ``Imagenet: A large-scale hierarchical image database,'' in \emph{2009 IEEE conference on computer vision and pattern recognition}, 2009, pp. 248--255.

\bibitem{oord2018representation}
A.~v.~d. Oord, Y.~Li, and O.~Vinyals, ``Representation learning with contrastive predictive coding,'' \emph{arXiv preprint arXiv:1807.03748}, 2018.

\bibitem{wang2021dense}
X.~Wang, R.~Zhang, C.~Shen, T.~Kong, and L.~Li, ``Dense contrastive learning for self-supervised visual pre-training,'' in \emph{Proceedings of the IEEE/CVF Conference on Computer Vision and Pattern Recognition}, 2021, pp. 3024--3033.

\bibitem{shin2023deep}
U.~Shin, J.~Park, and I.~S. Kweon, ``Deep depth estimation from thermal image,'' in \emph{Proceedings of the IEEE/CVF Conference on Computer Vision and Pattern Recognition}, 2023, pp. 1043--1053.

\bibitem{liu2021swin}
Z.~Liu, Y.~Lin, Y.~Cao, H.~Hu, Y.~Wei, Z.~Zhang, S.~Lin, and B.~Guo, ``Swin transformer: Hierarchical vision transformer using shifted windows,'' in \emph{Proceedings of the IEEE/CVF International Conference on Computer Vision}, 2021, pp. 10\,012--10\,022.

\bibitem{bian2021unsupervised}
J.-W. Bian, H.~Zhan, N.~Wang, Z.~Li, L.~Zhang, C.~Shen, M.-M. Cheng, and I.~Reid, ``Unsupervised scale-consistent depth learning from video,'' \emph{International Journal of Computer Vision}, vol. 129, no.~9, pp. 2548--2564, 2021.

\bibitem{loshchilov2018decoupled}
I.~Loshchilov and F.~Hutter, ``Decoupled weight decay regularization,'' in \emph{International Conference on Learning Representations}, 2018.

\bibitem{liu2016ssd}
W.~Liu, D.~Anguelov, D.~Erhan, C.~Szegedy, S.~Reed, C.-Y. Fu, and A.~C. Berg, ``Ssd: Single shot multibox detector,'' in \emph{European conference on computer vision}.\hskip 1em plus 0.5em minus 0.4em\relax Springer, 2016, pp. 21--37.

\bibitem{chen2021s2r}
X.~Chen, Y.~Wang, X.~Chen, and W.~Zeng, ``S2r-depthnet: Learning a generalizable depth-specific structural representation,'' in \emph{Proceedings of the IEEE/CVF Conference on Computer Vision and Pattern Recognition}, 2021, pp. 3034--3043.

\end{thebibliography}
}

\end{document}